\documentclass[10pt,twocolumn,letterpaper]{article}

\usepackage{cvpr}
\usepackage{times}
\usepackage{epsfig}
\usepackage{graphicx}
\usepackage{amsmath}
\usepackage{amssymb}
\usepackage{algorithm}
\usepackage[noend]{algpseudocode}
\usepackage{tabularx}
\usepackage{booktabs}
\usepackage{enumitem}

\usepackage[pagebackref=true,breaklinks=true,letterpaper=true,colorlinks,bookmarks=false]{hyperref}

\def\ie{\textit{i.e.}}

\def\etal{\textit{et al. \ }}
\newcommand{\pluseq}{\mathrel{+}=}

\cvprfinalcopy 

\begin{document}

\title{Online Adaptation through Meta-Learning for Stereo Depth Estimation}
\author{Zhenyu Zhang$^1$,
        ~St\'{e}phane~Lathuili\`{e}re$^2$,
        Andrea~Pilzer$^2$,
        ~Nicu~Sebe$^{2, 3}$,
        ~Elisa~Ricci$^{2, 4}$ 
        and~Jian Yang$^1$\\
	$^1$PCALab, Nanjing University of Science and Technology, China\\
	$^2$DISI, University of Trento, via Sommarive 14, Povo (TN), Italy\\
	$^3$Huawei Technologies Ireland, Dublin, Ireland\\
	$^4$Technologies of Vision, Fondazione Bruno Kessler, via Sommarive 18, Povo (TN), Italy\\
{\tt\small \{zhangjesse,csjyang\}@njust.edu.cn} \\
{\tt\small \{stephane.lathuiliere, andrea.pilzer,  niculae.sebe, e.ricci\}@unitn.it}\\
}

\maketitle

\begin{abstract}
In this work, we tackle the problem of online adaptation for stereo depth estimation, that consists in continuously adapting a deep network to a target video recorded in an environment different from that of the source training set. To address this problem, we propose a novel Online Meta-Learning model with Adaption (OMLA). Our proposal is based on two main contributions. First, to reduce the domain-shift between source and target feature distributions we introduce an online feature alignment procedure derived from Batch Normalization. Second, we devise a meta-learning approach that exploits feature alignment for faster convergence in an online learning setting. Additionally, we propose a meta-pre-training algorithm in order to obtain initial network weights on the source dataset which facilitate adaptation on future data streams.
Experimentally, we show that both OMLA and meta-pre-training help the model to adapt faster to a new environment. Our proposal is evaluated on the well-established KITTI dataset, where we show that our online method is competitive with state of the art algorithms trained in a batch setting. 
\end{abstract}

\section{Introduction}
\label{sec:intro}


Deep neural networks have brought amazing progresses in visual scene understanding in the last few years, enabling remarkable results in tasks such as object recognition \cite{he2016deep,krizhevsky2012imagenet}, semantic segmentation \cite{Zhang_2018_ECCV}, depth estimation \cite{eigen2015predicting} and many more. These advances can be ascribed not only to the availability of large scale datasets and powerful computational resources, but also to the design of specialized deep architectures. 

Depth estimation is one of the fundamental tasks in visual scene understanding and, over the years, has attracted considerable attention in the computer vision and robotics research communities. Earlier deep learning-based approaches for depth estimation considered a supervised setting: a deep regression model was trained to estimate a dense depth map from RGB images. This approach was exploited in many works~\cite{eigen2015predicting, laina2016deeper, fu2018dorn, liu2016learningtpami, xu2018structured, Yang2018ECCV, zhuo2015indoor} where it was shown that accurate depth maps can be recovered given enough training data. However, in the context of depth estimation, collecting the data is an expensive and time consuming task. For instance, in an autonomous driving setting it requires a car with a mounted camera system plus a LIDAR that drives for many hours in different environmental conditions. More recently, to avoid the costly procedure of collecting densely and accurately annotated datasets, researchers have proposed self-supervised, also known as unsupervised, depth estimation approaches.  
In the unsupervised setting a deep network is asked to regress the dense correspondence map (\ie \ disparity) between two views of a stereo image pair. Interestingly, recent works~\cite{garg2016unsupervised, godard2017unsupervised, zhou2017unsupervised, pilzer2018unsupervised} showed performance comparable to supervised methods on the common benchmark datasets (\textit{e.g.} KITTI~\cite{kitti2012,kitti2015}).

\begin{figure}[t]\centering
\includegraphics[width=0.85\linewidth]{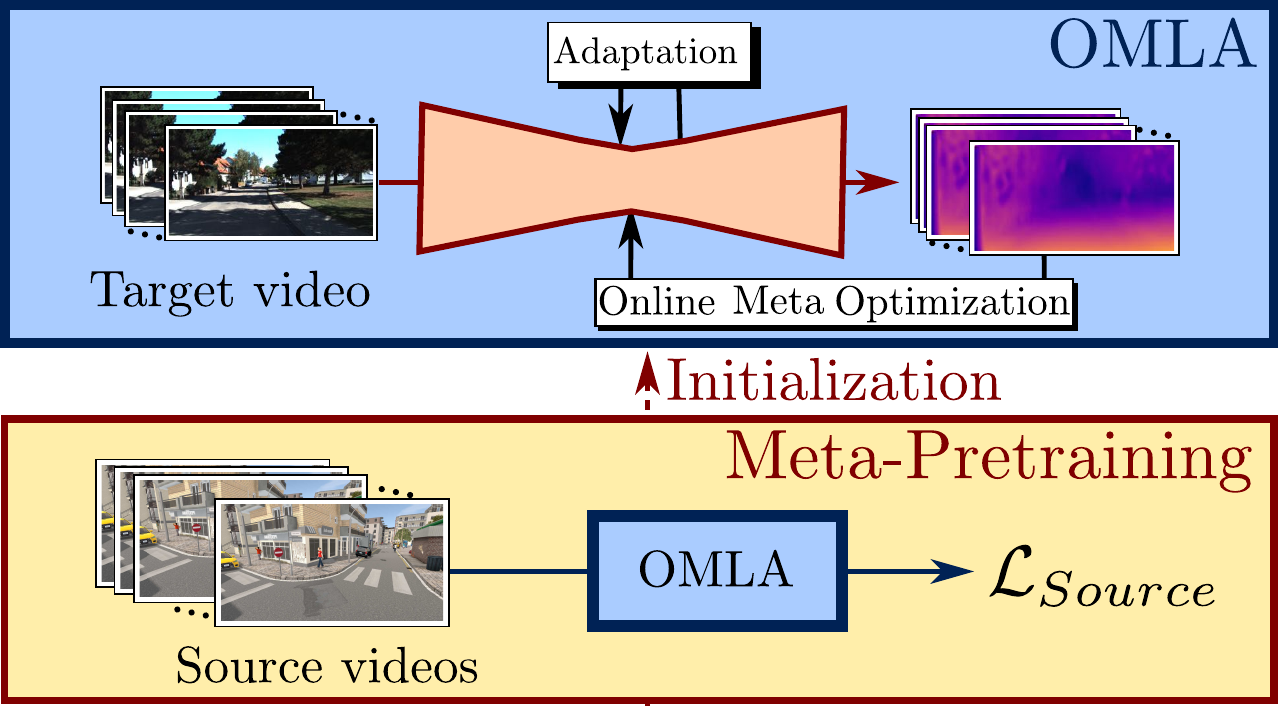}
\caption{We propose an Online Meta-Learning model with Adaption (\emph{OMLA}) for depth estimation. Our approach combines feature distribution alignment and meta-learning optimization for the purpose of predicting depth maps on a target video. 
Additionally, \emph{OMLA} is employed when pretraining on the source data in order to obtain a network whose parameters enable fast online adaptation. 
}
\vspace{-0.5cm}
\label{fig:teaser}
\end{figure}

One main limitation of current approaches is that they are designed and tested under a \textit{closed-world} assumption, meaning that training and testing data are derived from a common dataset and there is no much difference in term of visual appearance between video sequences used for learning the model and data considered for testing. 
In this paper we argue that, to be deployed in real applications (\ie \ for autonomous driving, robotics, etc), deep architectures for depth estimation should consider 
an \textit{open-world} setting, with visual data continuously gathered in changing environmental conditions. For example, in an autonomous driving scenario we would need a model that continuously adapt to changing environments (\ie \ urban, suburban, highway, etc) and lighting conditions (\ie \ night, dawn, day, tunnel, etc). In other words, we require a deep architecture with online adaptation abilities. 

Driven by this motivation, in this paper we propose a meta-learning approach for stereo depth estimation designed for fast online adaptation. Our proposal is illustrated in Fig.~\ref{fig:teaser}. First, we introduce an \emph{Online Meta-Learning Algorithm} (\emph{OMLA}) which combines feature distribution alignment and meta-learning (upper half of Fig.~\ref{fig:teaser}). Specifically, in order to handle the domain shift between the source training data and the target video, we first align the feature distributions of the two domains using statistics gathered along the video sequence and considering alignment layers derived from Batch Normalization similarly to \cite{mancini2018kitting,carlucci2017iccv}. Feature alignment is then combined with a meta-learning algorithm. The motivation for this choice is that, previous network parameter updates can be used to learn to update better on future frames. Since the frames of a video depict a similar environment, our meta-learning can learn how to optimize the network specifically for this environment.

In addition, we also propose to employ our \emph{OMLA} algorithm when pre-training the model on the source dataset (bottom half of Fig.~\ref{fig:teaser}), with the intent of obtaining a network parameter initialization that leads to accurate disparities after few frames of adaption only. {Rather than pre-training over stereo image pairs, we propose to explicitly use \emph{OMLA} on video sequences to define our training loss. Specifically, our meta-learning loss favors network parameters that lead to good depth predictions when using \emph{OMLA} on every video sequences of the source training set.}

To summarize our contributions are the following:
(i) We propose a novel approach for online adaptation in the context of depth prediction. Our method combines meta-learning and feature alignment 
to allow fast adaptation on video sequences recorded in new environments.
(ii) We introduce a meta-pre-training approach that explicitly uses \emph{OMLA} in order to provide a good parameter initialization for online adaption. (iii) From an experimental perspective, we perform an extensive evaluation on the well-known KITTI\cite{kitti2015} benchmark. We show that both our \emph{OMLA} and meta-pre-training help to improve depth prediction performance and that our method is even competitive with previous algorithms trained in a batch offline setting.

\section{Related Work}
\label{sec:related}

\textbf{Depth Estimation.} Depth estimation, among other scene understanding tasks, has attracted a lot of attention in the last years with the development of deep Convolutional neural Networks (ConvNets).
Deep models are usually trained following a supervised setting~\cite{eigen2015predicting, laina2016deeper, fu2018dorn, liu2016learningtpami, xu2018structured, Yang2018ECCV, zhuo2015indoor} by minimizing the discrepancy between the predicted and the ground truth depth maps. Eigen~\etal~\cite{eigen2015predicting} showed that a multi-scale approach leads to better performance and Laina~\etal~\cite{laina2016deeper} outlined the benefit of using a very deep architecture. Other works proposed to enforce some structure in the predicted depth maps considering graphical models as CRFs~\cite{liu2016learningtpami, xu2018structured, wang2015towards}. 
In order to train and evaluate these deep architectures, several datasets have been recorded for indoor scenes, as NYU~\cite{nyu}, or outdoor as KITTI~\cite{kitti2012, kitti2015} and Make3D~\cite{saxena2009make3d}.
Synthetic datasets, as Synthia~\cite{synthia2016}, have been also considered as an alternative in order to avoid the time consuming ground-truth recording process. However, the resulting models generally suffer from the domain shift between the synthetic and the real environments.

To avoid the need of annotated data, self-supervised depth estimation methods~\cite{garg2016unsupervised, godard2017unsupervised, zhou2017unsupervised, pilzer2018unsupervised} have been recently developed. 
For instance, 
Godard~\etal~\cite{godard2017unsupervised} used self-structuring and self-consistency losses to improve the prediction quality. Other works proposed to enhance the estimation accuracy through ego-motion estimation~\cite{zhou2017unsupervised}, adversarial learning~\cite{pilzer2018unsupervised, kundu2018cvpr}, visual odometry~\cite{Yang2018ECCV}.
Interestingly, Kundu \etal~\cite{kundu2018cvpr} tackled the problem of domain adaptation from synthetic to real world data proposing an adversarial approach for depth estimation. Recently, Tonioni \etal ~\cite{tonioni2019cvpr} employed a self-supervised formulation to enable fast update of parameters in an online setting for estimating depth maps. We follow this research direction and consider the problem of updating the prediction model online, but opposite to~\cite{tonioni2019cvpr}, we tackle explicitly the distribution misalignment problem and devise a novel strategy to obtain faster adaptation.

\begin{figure*}[t]\centering
\includegraphics[width=0.85\linewidth]{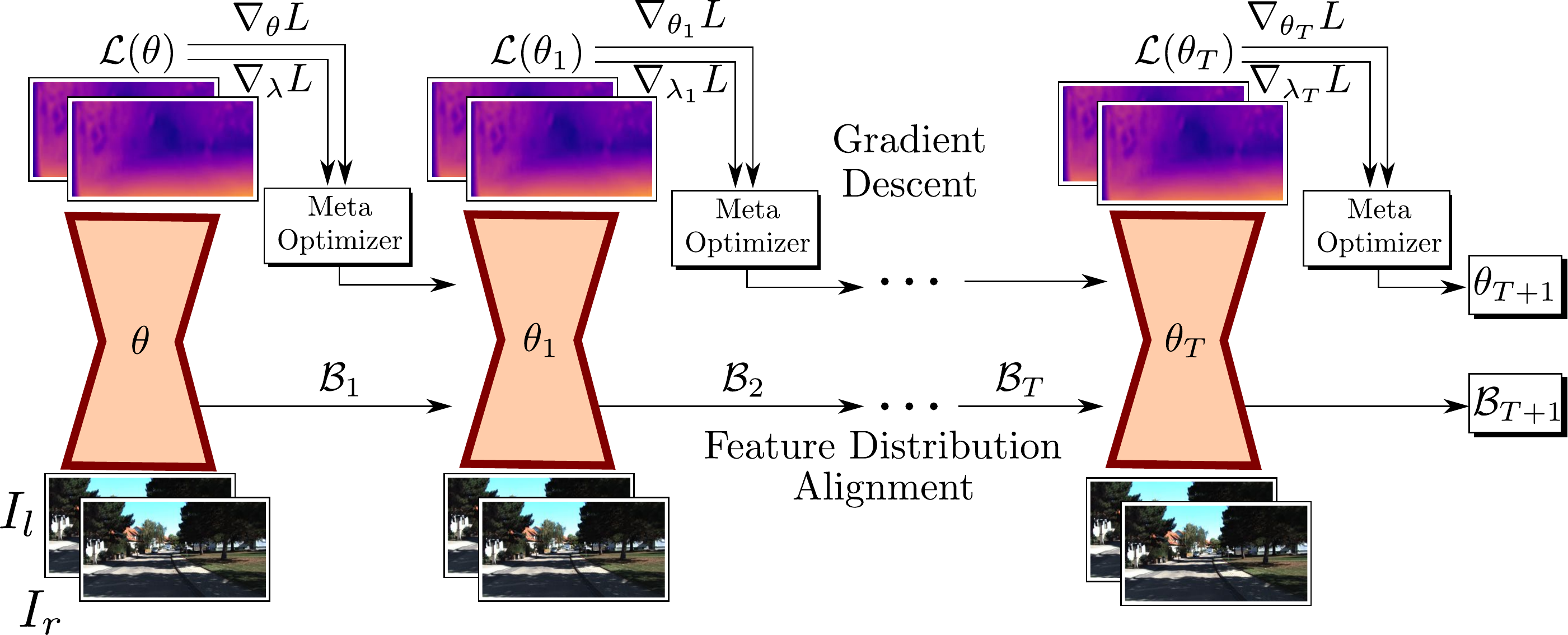}
\caption{Proposed Online Meta-Learning Algorithm (\emph{OMLA}). At time $t$, the feature $\mathcal{B}_t$ statistics are updated within BN layers for feature distribution alignment (Sec.~\ref{sec:adaBN}). Then, the model weights $\theta_t$ are updated according to our meta-learning optimizer (Sec.~\ref{sec:OMLA}).}
\vspace{-0.3cm}
\label{fig:pipeline}
\end{figure*}

\textbf{Domain Adaptation.}
Over the years, several works have considered the problem of domain adaptation within computer vision applications \cite{csurka2017domain}, proposing both shallow models and deep neural networks. Focusing on recent deep learning-based models, the different methods can be roughly grouped in three categories, according to the strategies used to reduce the discrepancy between the source and target feature distributions. The first category includes approaches which reduce the domain shift by designing appropriate loss functions, such as the Maximum Mean Discrepancy \cite{long2016deep,venkateswara2017deep} or the domain confusion loss \cite{tzeng2015simultaneous}. A second group of methods considers approaches based on Generative Adversarial Networks (GANs) \cite{Bousmalis:Google:CVPR17, Taigman2016UnsupervisedCI,sankaranarayanan2017generate}, whose main idea is to directly transform images from the target domain to the source domain. 
The latter category includes approaches which deal with the domain-shift problem by embedding into the deep architecture specifically designed {\em domain alignment layers}~\cite{li2016revisiting,carlucci2017autodial,mancini2018boosting}. The idea is to consider domain-specific Batch Normalization (BN) layers in order to align the source and target feature distributions to a common reference distribution. 
While most previous works on domain adaptation focused on a classification setting, recent works have considered structured predictions problems, such as semantic segmentation and depth prediction \cite{day2night2018, fog2018eccv}. 

Domain adaptation has been studied in the online learning setting where data are available sequentially and the target domain distribution changes continuously~\cite{hoffman2014cvpr, wulfmeier2018icra,mancini2018kitting}. Extending domain-alignment layers in \cite{li2016revisiting,mancini2018boosting}, online adaptation can be performed by incrementally updating feature statistics in the BN layers~\cite{mancini2018kitting}. 
In this work we consider domain adaptation in a pixel-level prediction problem, \textit{i.e.} depth estimation, and propose an elaborate formulation for online adaptation by combining feature distribution alignment through domain-specific layers and loss minimization. Furthermore, we introduce a meta-learning approach that improves the adaptation ability of the model trained on the source dataset.

\textbf{Meta-Learning.} Meta-learning is the problem of learning how to learn. In~\cite{vinyals2016matching, ravi2016optimization, finn2017model, li2018learning}, meta-learning has been employed to obtain fast generalization on novel domains or categories. In~\cite{ravi2016optimization,vinyals2016matching} the problem has been explicitly formulated as a few-shot learning problem.
 When it comes to deep network, meta-learning can improve convergence of gradient descent~\cite{andrychowicz2016learning} by using a trainable optimizer to train a neural network.
 For transfer learning applications, a \textit{policy network} has been proposed in~\cite{guo2018spottune} to decide which layer should be fine-tuned.
 Park~\etal~\cite{park2018meta} employed an offline meta-learning method to adjust the initial deep networks used in online tracking. 
Following this line of research, we consider meta-learning to obtain a source model that can adapt fast to a particular stereo video sequence. Conversely to \cite{park2018meta}, the use of feature distribution alignment is explicitly modeled in the source training algorithm. To the best of our knowledge, this is the first approach that introduces meta-learning in the context of online depth estimation.


\section{Meta-learning for Self-adaptive Depth Estimation}

In this section, we detail the proposed meta-learning approach for online adaptation.
Formally, we assume to have a source domain $\mathcal{S} =\{V_s^n\}_{n=1}^{N}$ composed of $N$ stereo video sequences recorded with the same calibrated stereo camera. In a first stage, we employ this source dataset to train a neural network $\Phi$ with parameters $\theta$ in order to predict the disparity maps between image pairs recorded with this stereo setting. The network training is performed via minimization of a loss $\mathcal{L}$ leading to network parameter values $\theta_0$. In a second stage, we consider that a target video sequence $V_{\mathcal{T}}$ is recorded using a different calibrated stereo camera in a different environment and that the video frame pairs $I_{\mathcal{T}}^t$ at time $t$ are available sequentially. The goal is to adapt the network parameters $\theta$ in order to predict more accurate disparity maps between image pairs recorded in this new environment. Note that, opposite to deep domain
adaptation in a batch setting \cite{kundu2018cvpr}, in this work we assume that the source dataset $\mathcal{S}$ is no longer available during this second stage.

A naive approach for online adaptation could consist in computing the training loss $\mathcal{L}$ on the current
frame and updating the whole network by gradient descent. This procedure could then be applied to each video frame. Despite its simplicity, this strategy has several drawbacks: it is very sensitive to domain shift, it accounts only for the current frame and it may introduce negative bias in the learning procedure. 
To cope with these issues, we propose to adapt our network by combining two complementary approaches: \emph{feature distribution alignment} and \emph{meta-learning} (see Fig.~\ref{fig:pipeline}). Specifically, to neutralize domain shift, we align the feature distributions using statistics gathered in the batch normalization layers and combined over time as detailed in Section~\ref{sec:adaBN}. By opposition to back-propagation-based loss minimization, feature distribution alignment is performed during the forward pass allowing adaption of the first layers with a limited computational cost. Additionally, we optimize our model with a fine-tuning strategy, and propose to guide the fine-tuning with a meta-learning optimizer (\textit{Meta-Optimizer} in Fig.~\ref{fig:pipeline}) for faster convergence as motivated and detailed in Section~\ref{sec:OMLA}. 
We argue that fine-tuning and feature distribution alignment are complementary to each other. Feature alignment can cope with low-level feature domain shifts, whereas fine-tuning can handle higher level representation shifts. 
In addition, we propose a meta-pre-training formulation to obtain initial parameters $\theta$ that are able to be adapted faster to a particular sequence. Our meta-pre-training strategy is explained in Section~\ref{sec:MT}. Finally, the whole model is trained using unsupervised depth estimation losses as detailed in Section~\ref{sec:Stereodepth}.

\subsection{Domain Adaptation via Online Feature Distribution Alignment (OFDA)}
\label{sec:adaBN}
We consider a deep network $\Phi$ embedding BN layers. We follow the idea of previous works \cite{carlucci2017iccv, li2016revisiting, mancini2018kitting} and perform domain adaptation by updating the BN statistics with the
incoming frames of the target video. The main idea behind this strategy is that the domain shift is reduced by aligning the target feature distribution to a gaussian reference distribution \cite{carlucci2017iccv, li2016revisiting, mancini2018kitting}.
For the sake of notation, here we consider a single BN layer but this approach is applied independently to each BN layer of $\Phi$. First, when training $\Phi$ on the source domain $\mathcal{S}$, we collect BN statistics $\mathcal{B}_s=(\mu_s,\sigma_s^2)$ as in \cite{li2016revisiting}. 
Second, we perform adaptation on the target video $V_{\mathcal{T}}$, using the following procedure. At time $t=0$, we initialize the batch statistics with $\mathcal{B}_o=\mathcal{B}_s$. At time $t$, we consider that we dispose of the previous BN statistics $\mathcal{B}_{t-1}=(\mu_{t-1},\sigma_{t-1}^2)$ at time $t-1$. Considering that we dispose of $m$ samples $\{x_1..x_m\}$ of a given feature vector, we compute the partial BN statistics:
\begin{align}
\hat{\mu}_t=\frac{1}{m}\sum_{i=1}^{m}x_i \hspace{0.5cm}  \hat{\sigma}_t^2=\frac{1}{m}\sum_{i=1}^{m}(x_i-\hat{\mu}_t)^2 
\end{align}
Given a dynamic parameter $a \in \mathbb{R}$, the global statistics are computed as follows:
\begin{align}
  &\hat{\mu}_t= (1-a) \mu_{t-1}+  a\hat{\mu}_t\notag \\
  &\sigma_t^2=(1-a)\sigma_{t-1}^2+  a\frac{m}{m-1}\hat{\sigma}_{t}^2 \notag
\end{align}
For a given input $x$, the output of the normalization layer is then given by:
\begin{align}
\hat{x}=\gamma \frac{x-\mu_t}{\sqrt{\sigma_t^2+\epsilon}}+\beta
\end{align}
where $\gamma$ and $\beta$ are the usual affine transformation parameters of the BN layer, while $\epsilon\in\mathbb{R}$ is a
constant introduced for numerical stability.

\subsection{Online Meta-learning with Adaptation (\emph{OMLA})}

\label{sec:OMLA}


We now introduce our online meta-learning approach with adaptation.
We assume to have a target stereo video sequence $V=\{I_t\}_{t=1}^T$ of length $T$. When performing online training on $V$, we use the following recursive algorithm given in Alg.~\ref{alg:OMLA}.
\begin{algorithm}
\caption{Online Meta-Learning with Adaptation}
\label{alg:OMLA}
\begin{algorithmic}[1]

\Procedure{OMLA}{$V=\{I_t\}_{t=1}^T,\theta_0,\mathcal{B}_{0},\lambda_0,\lambda$}       
\For{t=0..T}
\State $D_t,\mathcal{B}_{t+1}=\Phi\left((\theta_t,\mathcal{B}_{t}),I_t\right)$
\State $\mathcal{L}_t=\mathcal{L}(D_t,I_t)$
\If{$t> 0$}
\State $\lambda_{t}=\text{Optimizer}(\nabla_{\lambda_{t-1}}\mathcal{L}_t,\lambda)$
\EndIf
\State $\theta_{t+1}=\text{Optimizer}(\nabla_{\theta_t}\mathcal{L}_t,\lambda_{t})$
\EndFor
\Return $\theta_{t+1},\mathcal{B}_{t+1}$
\EndProcedure
\end{algorithmic}

\end{algorithm}
For initialization, we assume we dispose of network parameters $\theta_{0}$, BN statistics $\mathcal{B}_0$, an initial learning rates for the network parameters $\lambda_0$ and a meta-learning rate $\lambda$. Note that, we use specific learning rates for each network parameter. Therefore, both $\lambda_0$ and $\lambda$ have the same dimension as $\theta$.
At time $t$, given the current parameters $\theta_t$, the disparity maps between the stereo pair $I_t$ are predicted according to $D_t=(d_r,d_l)=\Phi((\theta_t,\mathcal{B}_{t}),I_t)$. Here, $d_l$ and $d_r$ denote the left and right disparities. Note that, in the forward pass, we perform feature distribution alignment using statistics gathered in the BN layers as described in Sec.~\ref{sec:adaBN}. The BN statistics $\mathcal{B}_{t+1}$ are stored for the next iteration.
The predicted disparity quality is assessed via $\mathcal{L}(D_t,I_t)$ where $\mathcal{L}$ is a loss function detailed in \ref{sec:Stereodepth}. Then, we udpate the network parameter learning rate by performing one gradient descent in order to minimize $\mathcal{L}$ with respect to the learning rate at the previous \emph{OMLA} iteration $\lambda_{t-1}$. The motivation here is to obtain better learning rates for the next network parameter updates. The gradient descent step can be computed using any gradient-based optimizer. In all of our experiment, we employ the Adam optimizer~\cite{kingma2014adam}.
Finally, we update the network parameter by applying a gradient descent step. The procedure returns the final network parameter $\theta_{T+1}$ and the BN statistics $\mathcal{B}_{T+1}$. In the case of online learning on the target video, the parameters obtained at the end of \emph{OMLA}, $\theta_{T+1}$ and $\mathcal{B}_{T+1}$, are not further used. Nevertheless, they are used when \emph{OMLA} is employed within our meta-pre-training procedure described in the next section.

\subsection{Meta-pre-training for Fast Adaptation}
\label{sec:MT}
In this section, we detail our meta-learning framework for pre-training our network on the source dataset.
The motivation behind our meta-learning formulation is to obtain network parameters that lead to accurate disparity predictions on image pairs of our source dataset but also that can be adapted to a specific sequence in a few frames only. Rather than using a training procedure that would minimize a reconstruction loss over stereo pairs, we propose to explicitly use \emph{OMLA} to define our training loss. 
Consequently, our loss enforces that the network parameters must lead to accurate depth predictions after using \emph{OMLA} on every video sequence of the source training set.
Formally, we assume to have a source domain $\mathcal{S} =\big\{\{I_t^k\}_{t=1}^T\big\}_{n=1}^{N}$ composed of $N$ video sequences of length $T$ recorded with the same calibrated stereo camera. We consider here video sequences of equal lengths $T$ for the sake of notation but it could be applied to any arbitrary varying lengths.
We seek network parameters $\theta$ that lead to a low loss value after N steps of \emph{OMLA}, $\mathcal{L}(\Phi(D,\theta_T),I_T)$. Importantly, $\theta$ must lead to a fast adaptation given that feature distribution alignment is employed when learning online. In addition, we propose to use the meta-learning pre-training procedure to provide \emph{OMLA} hyper-parameters such as the learning rates $\lambda$. Interestingly, the meta-learned learning rates $\lambda$ can be interpreted as a hyper-parameter indicating which network parameters should be fine-tuned and which parameters should not be updated. Our meta-training procedure consists in repeating the training step given in Alg.\ref{alg:MT} until convergence. 

\begin{algorithm}[t]
  \caption{Meta-Training Step for adaptation}
  \label{alg:MT}
\begin{algorithmic}[1]

\Procedure{step}{$\big\{\{I_t^k\}_{t=1}^T\big\}_{k=1}^{K},\theta,\mathcal{B},\lambda,\lambda_{\theta},\lambda_{\lambda}$}       
\State $\text{grad}_{\theta}=\text{grad}_{\lambda}=0$
\For{k=1..K} \Comment{For each video of the meta-batch}
\State $\theta_N^k,\mathcal{B}_N^k=\text{OMLA}(\{I_t^k\}_{t=1}^{N},\theta,\mathcal{B},\lambda,\lambda_\lambda)$
\For{$t=N\!+1..T$} \Comment{Evaluation of $\theta_N^k,\mathcal{B}_N^k$}
\State $D_t=\Phi\left((\theta_N^k,\mathcal{B}_N^k),I_t^k\right)$
\State $\mathcal{L}_t=\mathcal{L}(D_t,I_t)$
\State $\text{grad}_{\theta}\pluseq\nabla_{\theta}\mathcal{L}_t$
\State $\text{grad}_{\lambda}\pluseq\nabla_{\lambda}\mathcal{L}_t$
\EndFor
\EndFor
\State $\lambda= \text{Optimizer}(\text{grad}_{\lambda},\lambda_\lambda)$ \Comment{Updates}
\State $\theta=\text{Optimizer}(\text{grad}_{\theta},\lambda_{\theta})$

\Return $(\theta,\lambda)$
\EndProcedure
\end{algorithmic}
\end{algorithm}

The procedure takes as inputs a subset of the source dataset $\big\{\{I_t^k\}_{t=1}^T\big\}_{k=1}^{K}\subset\mathcal{S}$ composed of $K$ videos. These $K$ videos form a meta-batch containing K different cases where the network is adapted to a particular video. We provide also, the current network parameters $\theta$ together with BN Statistics $\mathcal{B}$. Finally, we provide three different learning rates:
the current meta-learned learning rate $\lambda$ and the two fixed learning rate $\lambda_{\theta}$ and $\lambda_{\lambda}$ for the network parameters and the meta-learning rate respectively. For each video of the meta-batch, our algorithm is divided in two steps. First, we employ \emph{OMLA} on the first $N$ frame pairs to adapt specifically to the video. For the $k^{th}$ video, we obtain network parameters $\theta_N^k$ and BN statistics $\mathcal{B}_N^k$. The second step consists in evaluating the parameters obtained with \emph{OMLA} on the remaining frames of the video.
We compute the loss function and its gradient with respect to the original $\theta$ parameters used as initialization of the \emph{OMLA}. The motivation for computing this gradient is that we aim at obtaining $\theta$ parameters that leads to fast adaptation and therefore low loss values $\mathcal{L}_t$. The gradients are summed over all the future frames of the $K$ videos of the meta-batch. The same procedure is applied for the learning rate $\lambda$. Finally, we perform two gradient descent steps using the computed gradients and an \textit{ad hoc} optimizer for $\theta$ and $\lambda$ respectively. 

\begin{table*}[t]
\begin{center}
\resizebox{.92\textwidth}{!}{
\begin{tabular}{ | l | c|| c | c | c | c || c | c | c | c |}
\toprule
&&\multicolumn{4} {c||} {Online Evaluation Scores}&\multicolumn{4} {c|} {Evaluation Scores on Last 20\% frames}\\
   Method &Pre-training& RMSE & Abs Rel & Sq Rel & $RMSE_{log}$ & RMSE & Abs Rel & Sq Rel & $RMSE_{log}$\\
   \midrule
Online Naive & w/o & 12.2012&0.4357&5.5672&1.3598&12.2874&0.4452&5.5213&1.3426\\
  \hline
   Online Naive& Standard &9.0518 &0.2499 &3.2901 &0.9503 &9.0309 &0.2512 &3.3104 &0.9495 \\
   Online Meta-learning& Standard&8.7553 &0.2367 &3.0028 &0.9412 &8.7032 &0.2285 &2.9842 &0.9403 \\
   OFDA& Standard&4.7280 &0.1885 &1.3012 &0.2331 &4.6134 &0.1800 &1.2957 &0.2297 \\
   OMLA& Standard&4.5126 &0.1623 &1.2892 &0.2287 &4.4783 &0.1503 &1.2033 &0.2198 \\
  \hline
 Online Naive& Meta &8.8230  &0.2305  &3.0578  &0.9324  &8.7061  &0.2273  &2.9804  &0.9065 \\
 Online Meta-learning & Meta&8.5572  &0.2301  &2.9576  &0.9054  &8.4325  &0.2278  &2.8503  &0.8921 \\
 OFDA  & Meta &4.1279  &0.1236  &0.9027  &0.1989  &4.0731  &0.1176  &0.8845  &0.1921 \\ 
 OMLA & Meta &\bf 3.9025  &\bf 0.1189  &\textbf{0.8256}  &\bf 0.1952  &\bf 3.7203  &\bf 0.1058  &\textbf{0.8176}  &\textbf{0.1835} \\ 
\bottomrule
\end{tabular}
}
\end{center}
\caption{Ablation study on KITTI Eigen test split of the proposed unsupervised online \textbf{stereo} method. At the top we show fine-tuning without pre-training, in the middle part fine-tuning after standard batch pre-training and in the bottom part fine-tuning after meta-pre-training on Synthia dataset. 
Depth predictions are capped at 50 meters. 
}
\vspace{-5pt}
\label{stereo_result}

\end{table*}

\subsection{Depth Estimation Loss}
\label{sec:Stereodepth}
In this section, we provide the loss $\mathcal{L}$ used both in \emph{OMLA} and in our meta-pre-training algorithm (see $\mathcal{L}$ in Algs.\ref{alg:OMLA} and \ref{alg:MT}). Following \cite{godard2017unsupervised}, the model takes in input the left image $I_l$ and the right image $I_r$ and outputs the disparities $d_l, d_r$. We employ a warping operation $f_w$ in order to reconstruct the left image from the right image according to:
\begin{equation}
    \hat{I}_l = f_w(I_r, d_l)
\end{equation}
Symmetrically, we obtained a reconstructed right image $\hat{I}_r$ form the left image.
The loss is a combination of a reconstruction loss $\mathcal{L}_1$ and a self-structuring loss (SSIM) $\mathcal{L}_{SSIM}$ proposed in~\cite{godard2017unsupervised} weighted by a parameter $\alpha=0.85$
\begin{equation}
\begin{split}
\mathcal{L} = (1-\alpha)|| \hat{I}_l - I_l ||_1 + \alpha\mathcal{L}_{SSIM}(\hat{I}_l, I_l) \\+ (1-\alpha)|| \hat{I}_r - I_r ||_1 + \alpha\mathcal{L}_{SSIM}(\hat{I}_r, I_r)
\end{split}
\end{equation}
Using such reconstructing loss, we can perform depth estimation in a totally unsupervised way, and perform adaptation in an online mode without groud truth.

\section{Experiments}
\label{sec:exp}

\begin{table*}[!t]
  \begin{center}
      \resizebox{.92\textwidth}{!}{
    \begin{tabular}{|l||c|c|c|c||c|c|c|c||c|}
\toprule
&\multicolumn{4} {c||} {Pretraining on Synthia~\cite{synthia2016}}&\multicolumn{4} {c||} {Pretraining on SceneFlow~\cite{mayer2016large}}&\\
  Method& RMSE & Abs Rel & Sq Rel & $RMSE_{log}$ & RMSE & Abs Rel & Sq Rel & $RMSE_{log}$ & FPS \\
  \midrule
  \hline
  DispNet~\cite{mayer2016large} Naive& 9.0222 & 0.2710 & 4.3281 &0.9452 &9.1587 &0.2805 &4.3590 &0.9528 & \bf 5.42 \\
  DispNet ours & \bf 4.5201 & \bf 0.2396 & \bf 1.3104 & \bf 0.2503 &\bf 4.6314& \bf 0.2457&\bf 1.3541&\bf 0.2516 & 4.00 \\
  \hline
  MADNet~\cite{tonioni2019cvpr} Naive &  8.8650 & 0.2684 &  3.1503&0.8233&8.9823&0.2790&3.3021&0.8350 & \bf 12.05 \\
  MADNet ours & \bf 4.0236 & \bf 0.1756 & \bf 1.1825 &\bf 0.2501 & \bf 4.2179 &\bf 0.1883 &\bf 1.2761 &\bf 0.2523 & 9.56 \\

  \hline
  Godard  \etal (ResNet)~\cite{godard2017unsupervised} Naive & 9.0518 & 0.2499 & 3.2901 &0.8577 & 9.0893 & 0.2602 & 3.3.896 & 0.8901 & \bf 5.06
  \\Godard  \etal (ResNet) ours & \bf 3.9025 & \bf 0.1189 & \bf 0.8256 &\bf 0.1952 & \bf 4.0573 & \bf 0.1231&\bf 1.1532&\bf 0.1985& 3.40 \\

\hline
    \end{tabular} 
}
    \end{center}
\caption{Analysis of the performance of our method on common stereo architectures, \textit{DispNet}~\cite{mayer2016large}, \textit{MADNet}~\cite{tonioni2019cvpr} and~\textit{Godard} (ResNet)~\cite{godard2017unsupervised}, and different pretraining datasets, Synthia~\cite{synthia2016} or SceneFlow~\cite{mayer2016large}. Depth predictions are capped at 50 meters.}
  \label{net_archi} 
\end{table*}

\begin{table*}[t]

\begin{center}
\resizebox{.85\textwidth}{!}{
\begin{tabular}{ | l  || c | c | c | c || c | c | c |}
\toprule
   Method  & RMSE & Abs Rel & Sq Rel & $RMSE_{log}$ & $\alpha>1.25$ & $\alpha>1.25^2$ & $\alpha>1.25^3$\\
   \midrule
Godard \etal \cite{godard2017unsupervised} Offline  & \textbf{3.6975} & 0.0983 & 1.1720& \textbf{0.1923}& 0.9166 & 0.9580 & \textbf{0.9778} \\ 
Godard \etal \cite{godard2017unsupervised} Offline + Online & 3.7059 & \textbf{0.0980}& 1.1712& 0.1956& \textbf{0.9203} & \textbf{0.9612} & 0.9776 \\
Godard \etal + \emph{OMLA}& 3.9025  &0.1189  &\textbf{0.8256}  &0.1952  &0.9110  &0.9505  &0.9776 \\
\hline
MADNet~\cite{tonioni2019cvpr} Offline  & \textbf{3.8965}& 0.1793   &1.2369 &  \textbf{0.2457}  & 0.9147 & 0.9601  & 0.9790 \\
MADNet~\cite{tonioni2019cvpr} Offline + Online & 3.9023 & 0.1760 & 1.1902 & 0.2469 &\bf 0.9233 &\bf 0.9652& \bf 0.9813 \\
MADNet + \emph{OMLA} & 4.0236 &\textbf{0.1756} &\textbf{1.1825} &0.2501 &0.9022 &0.9453 &0.9586 \\
\hline
DispNet~\cite{mayer2016large} Offline  & 4.5210 & 0.2433 & \textbf{1.2801}& \textbf{0.2490} & 0.9126 & 0.9472 & \textbf{0.9730} \\ 
DispNet~\cite{mayer2016large} Offline + Online & 4.5327 & \bf 0.2368 & 1.2853 & 0.2506 & \bf 0.9178 & \bf 0.9600 & 0.9725 \\
DispNet + \emph{OMLA} & \textbf{4.5201} & 0.2396 & 1.3104 &0.2503 &0.9085 &0.9460 &0.9613\\

\bottomrule
\end{tabular}
}
\end{center}
\caption{
Comparison with different offline methods. Only points with depth less than 50m are calculated. 
}
\label{offline_compare}
\end{table*}

\subsection{Evaluation}\label{evaluation}
\textbf{Evaluation for online learning.}
Following an online learning protocol, the frames are fed into the network sequentially. Each frame leads to a predicted depth map and a model parameter update. Importantly, we evaluate the estimated depth maps obtained at each time step before applying gradient descent. After processing the whole sequence, we compute the average scores over the sequence. In order to further evaluate the adaptation ability of the different models, we also report the average scores over the last 20\% frames of each video. The motivation behind these scores is that they measure the final prediction quality after convergence whereas the average scores over the whole sequence measure better the convergence speed.


\noindent
\textbf{Evaluation metrics.}
The quantitative evaluation is performed according several standard metrics used in previous works~\cite{eigen2015predicting, godard2017unsupervised, wang2015towards}. Let $P$ be the total number of pixels in the test set and $\hat{d}_i$, $d_i$ the estimated depth and ground truth depth values for pixel $i$. We compute the following metrics: 
\vspace{-0.5cm}
\begin{itemize}[noitemsep]\setlength
\item  Mean relative error (abs rel): 
$\frac{1}{P} \sum_{i=1}^{P} \frac{\parallel \hat{d}_i - d_i \parallel}{d_i}$, 
\item Squared relative error (sq rel): 
$\frac{1}{P} \sum_{i=1}^{P} \frac{\parallel \hat{d}_i - d_i \parallel^2}{d_i}$, 
\item Root mean squared error (rmse): 
$\sqrt{\frac{1}{P}\sum_{i=1}^P(\hat{d}_i - d_i)^2}$, 
\item Mean $\log10$ error (rmse log):
$\sqrt{\frac{1}{P} \sum_{i=1}^{P} \parallel \log \hat{d}_i - \log d_i \parallel^2}$
\item Accuracy with threshold $\tau$, \ie~the percentage of $\hat{d}_i$ such that $\delta = \max (\frac{d_i}{\hat{d}_i},\frac{\hat{d}_i}{d_i}) < \alpha^\tau$. We employ $\alpha = 1.25$ and $\tau \in [1,2,3]$ following \cite{eigen2015predicting}.
\end{itemize}
 
\noindent
\textbf{Datasets.}
Evaluation of adaptation methods requires two different datasets: a source and a target dataset. As source dataset, we select synthetic datasets which contain videos of driving environment. To evaluate the online adaptation performance, we select a real-world urban dataset that is used as the target dataset. In detail, we use the following benchmarks:

\noindent
\textit{Synthia dataset:} Synthia~\cite{synthia2016} is a synthetic dataset made of urban driving scenes. It contains stereo image pairs for four views, frontal, rear, left and right views. There are five video sequences for each of the four seasons, spring, summer, fall and winter. We select 4k frontal view image paris from the spring recordings, and use these images as source dataset to perform our meta-pretraining procedure.

\noindent
\textit{Scene Flow Driving:} Scene Flow Driving~\cite{mayer2016large} is a synthetic dataset with one driving video. It contains different settings on camera, speed and direction. We select all 2k stereo image pairs in forward setting for meta-pretraining.

\noindent
\textit{KITTI:} As target domain, we employ the~KITTI~\cite{kitti2015} dataset. KITTI is recorded from driving vehicles. We employ the training and test split of Eigen \etal~\cite{eigen2015predicting}. This split is composed of 32 different scenes for training, and 28 different driving scenes for testing. Note that, for online evaluation, we use all stereo images in the testing sequences.

\noindent
\textbf{Implementation Details.}
We implement the proposed method using Pytorch~\cite{paszke2017automatic} on a single Nvidia P40 GPU. All of the networks we built contain batch normalization layer \cite{ioffe2015batch} to perform the proposed feature distribution alignment. For pretraining on each synthetic dataset, we first perform unsupervised learning for a total 200 epochs in an offline batch setting, with initial learning rate $1e-4$ for 100 epochs, halved to $5e-5$ for the remaining 100 epochs. Then we perform meta-pretraining as Alg.~\ref{alg:MT} for 10 epochs. Here we set $\lambda = 1e-4, \lambda_\lambda = 1e-5, \lambda_\theta = 1e-5$, meta-batch $K = 8$ and future frames $T=3$. For online learning, we perform adaptation following Alg.~\ref{alg:OMLA} and set the meta-learning rate $\lambda = 1e-7$. All of the networks are trained from scratch with the Adam optimizer~\cite{kingma2014adam}.

\subsection{Results and Analysis}
In this section, we evaluate our proposed approach, experimentally validate the benefit of each component, and compare its performance with state-of-the-art methods. 

\noindent
\textbf{Analysis on the Proposed Method.}
To validate the contribution of each component of our method, we adopt the framework proposed in \cite{godard2017unsupervised} and used in several recent works \cite{pilzer2018unsupervised,zhou2017unsupervised,Yang2018ECCV}. 
For fair comparison, all the online learning procedures are applied and evaluated on the videos of Eigen's testing split \cite{eigen2015predicting} and, except explicit specification, all models are 
pretrained on the Synthia dataset. 

As a first baseline for online learning, we consider the approach that consists in performing adaptation via gradient descent at every step using a fixed learning rate. This approach is referred to as \emph{Online Naive}. Note that \emph{Online Naive} is equivalent to our approach without feature distribution alignment and without meta-learning updates in Alg.\ref{alg:OMLA}. We consider three different variants of our model. First, in \emph{Online Meta-learning}, we employ meta-learned gradient updates in Alg.\ref{alg:OMLA} but we do not use feature distribution alignment. Second, in  \emph{OFDA}, we use feature distribution alignment but perform gradient descent as in \emph{Online Naive}. Finally, in \emph{OMLA}, we use our full model. 

Concerning the pretraining, we compare different models where we use either batch pretraining (referred to as \textit{Standard} in upper half) or our meta-pretraining (referred to as \textit{Meta} in bottom half). For completeness, we also report the performance of a model without pretraining. 

We report the evaluation scores obtained by the different methods, in Table \ref{stereo_result}.
 First, we observe that directly performing naive online learning without pretraining (\textit{Online Naive, w/o}) does not lead to good performance. The scores obtained on the last $20\%$ frames are not better than the average scores over the whole videos showing that the model is not learning.
Similarly, moving to the models pretrained with \textit{Standard} pretraining on Synthia, \textit{Online Naive} only provides a very limited gain. A first proof that online meta-learning is beneficial is found in \textit{Online Meta-learning} where we see a clear improvement in the last $20\%$ frames when applying meta-learning for online fine-tuning. Even better performance are obtained with \textit{OFDA} that perform feature distribution alignment. These results show that handling domain shift between the source and the target distributions truly improve the quality of the estimated depth maps. A further improvement is obtained with our full meta-learning method, \emph{OMLA}, that reaches the best performances in the \textit{Standard} pretraining setting.

Concerning pretraining strategy, our meta-pretraining approach, denoted by \textit{Meta}, improves consistently the performances for every setting adopted on the target video with respect to \textit{Standard}. Similarly to what is observed with standard pretraining, \textit{Online Naive} does not perform well. \textit{Online Meta-learning} with meta-pretraining obtains better results considering both the scores averaged over the complete sequences and over the last 20\% frames. This indicates meta-pretraining helps the model, not only to adapt faster, but also to perform better after observing many frames. Furthermore, \textit{OFDA} 
 again improves the performance. The gain is even larger on the last $20\%$ of the frames. Finally, \emph{OMLA} achieves the best performance by combining \emph{OFDA} with meta-learning. 




\noindent
\textbf{Analysis on Network Architectures and Datasets.}
In order to further evaluate our approach, in Table~\ref{net_archi} we report the performances of our method considering three different architectures:  DispNet~\cite{mayer2016large}, MADNet~\cite{tonioni2019cvpr} and Godard \etal (ResNet encoder)~\cite{godard2017unsupervised}. 
DispNet and MADNet are two light-weight networks for stereo matching. We compare the performances obtained using these network architectures when we employ the baseline \textit{naive} online learning approach and our full model referred to as \textit{ours}. We report results when pretraining either on Synthia or SceneFlow Driving datasets.

From Table~\ref{net_archi}, we see that our proposed method obtains significantly better performances independently of the architecture. Such results demonstrate that \emph{OMLA} is effective even with smaller networks as DispNet or MADNet. 
Concerning the SceneFlow driving dataset, we observe that the models, both \emph{Naive} and ours, obtain slightly poorer performance than when pretraining on Synthia. A possible explanation is that the SceneFlow driving dataset is smaller and less diverse than Synthia. Nonetheless, these experiments confirm again the excellent performance of our approach even on this small dataset.

\begin{figure*}[t]
    \centering
    \includegraphics[width=.98\textwidth]{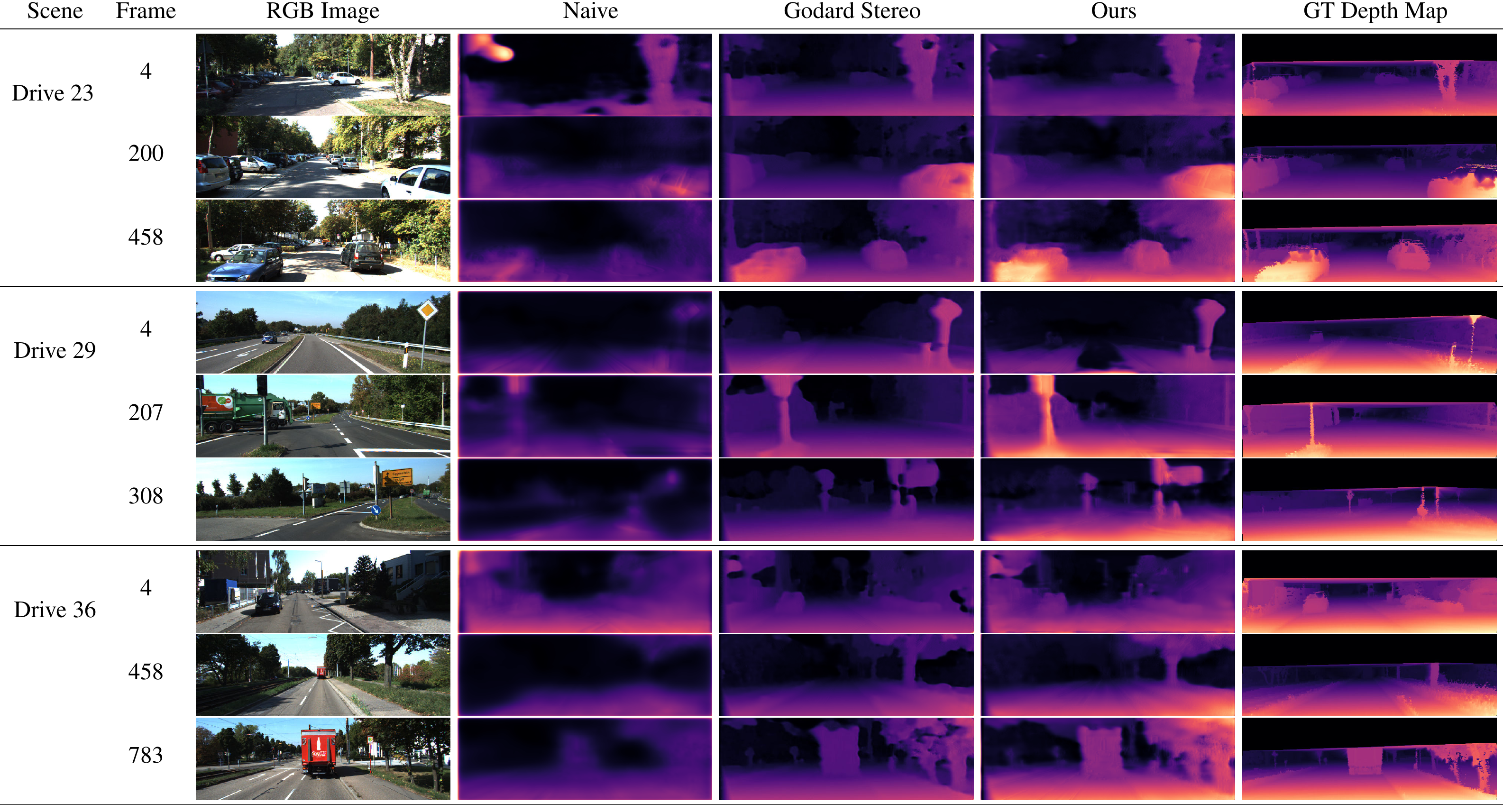}
    \caption{Qualitative comparison of different baseline models of the proposed approach on three video sequences of the KITTI dataset Eigen test split. We report frames at the beginning, in the middle and at the end of each video.}
    \label{fig:qualitative}
\end{figure*}

Concerning running time, the reported \textit{frames per second} (FPS) are reduced by approximately $20\%$ since \emph{OMLA} requires more gradient computation and parameter updates. Nevertheless, taking into account the performance gains we claim that such running time increase is acceptable for most applications.

\subsection{Comparison with offline methods}
In this section we compare our online learning method with models trained in an offline setting.  We consider the following baselines:
\vspace{-0.1cm}
\begin{itemize}[noitemsep]\setlength
\item \textit{Offline}, model pretrained using offline training as in \cite{godard2017unsupervised} on the KITTI Eigen training split and tested on KITTI Eigen test split. 
\item \textit{Offline+Online}, model pretrained using standard offline training on KITTI Eigen training split and online learning on KITTI Eigen test split. In that case, we employed the \emph{naive} online formulation previously described.
\item \emph{OMLA}, model meta-pretrained offline on Synthia and using \emph{OMLA} on KITTI Eigen test split. \end{itemize}
\vspace{-0.1cm}
The results are reported in Table~\ref{offline_compare}. First, when the models are trained in an \textit{Offline} setting on the KITTI training set, \emph{naive} online learning does not improve significantly the performance. 
Second, we observe that our online approach is competitive with the methods trained offline on the KITTI training set whereas our model did not see any real-world KITTI image. According to some metrics, our approach even outperforms the model trained offline. These observations clearly show the potential of our approach. 

Finally, we report qualitative results in Fig.~\ref{fig:qualitative}. We show the input frames, and the associated predictions, from the beginning, the middle and the last part of a same video. In the first frame of the video, we observe that the offline method already performs well, while the \emph{naive online} model and our model obtain poor results. The reason is that these two models were trained on totally different environments and did not observe enough frames to adapt. However, after several frames, we see that our method is able to learn and improve its predictions, while the naive model improves its performance more slowly. Finally, after observing enough frames, our model produces satisfactory results getting closer to the offline model predictions and to the ground truth. These qualitative results demonstrate that our method adapts effectively in a new environment and progressively improves its estimations.

\section{Conclusions}
We addressed the problem of online domain adaptation in the context of depth estimation and presented an algorithm, \emph{OMLA}, specifically designed for a sequential learning setting where fast convergence of the network training and adaptation to evolving data streams are required. 
We evaluated the proposed framework on the challenging KITTI dataset where we achieved state-of-the-art performance. As future works, we plan to combine our approach with the fast network update method in \cite{tonioni2019cvpr} and to extend our framework in a monocular setting.

{\small
\bibliographystyle{ieee}
\bibliography{egbib}
}

\clearpage
\appendix
\section*{Appendix}
We now report additional experiments using a monocular setting for depth estimation in order to further compare our approach with \cite{godard2017unsupervised} (see Sec.~\ref{sec:mono}). Then, in Sec.~\ref{sec:SMmetrics}, we evaluate our approach according to stereo matching metrics. Finally, in Sec.~\ref{sec:illu}, we study the temporal behavior of several variants of our proposed model. 

\section{Analyses on monocular setting}
\label{sec:mono}
Although our method is meant for online stereo depth estimation, we also report experiment results in a monocular setting for further evaluation. More precisely, we employ the monocular network of \cite{godard2017unsupervised} but still employed stereo pairs to compute the loss as in \cite{godard2017unsupervised}. 
The results obtained in the monocular setting are reported in Table \ref{mono_result}. As in the main paper, we also show the results averaged over the last 20\% frames of each scene. We can observe that directly performing naive online learning without pretraining does not lead to good performances. We notice that the scores on the last $20\%$ frames are not better than the scores averaged over the whole video showing that the model is not learning.
Concerning the online methods with pretraining, the results are well in-line with the stereo setting results reported in the main paper. We first observe that our meta-pretraining strategy improves consistently the performance for every strategy adopted on the target video. With meta-pretraining, the models can all obtain better results on the last 20\% frames, which show again that meta-pretraining helps the model, not only to adapt faster, but also to perform better after observing many frames. Finally, using Online Feature Distribution Alignment (OFDA) and online meta-learning both improve the performances of online learning. Similarly to the stereo setting, OMLA (online meta-learning with OFDA) leads to the best results with both standard and meta pretraining. 

\begin{table*}[h]
\centering
\resizebox{.99\textwidth}{!}{
\begin{tabular}{ | l | c|| c | c | c | c || c | c | c | c |}
\toprule
&&\multicolumn{4} {c||} {Online Evaluation Scores}&\multicolumn{4} {c|} {Evaluation Scores on Last 20\% frames}\\
   Method &pretraining& RMSE & Abs Rel & Sq Rel & $RMSE_{log}$ & RMSE & Abs Rel & Sq Rel & $RMSE_{log}$\\
   \midrule
Naive Online FT \cite{godard2017unsupervised} & w/o  & 13.4035 & 0.4687 & 5.7436 & 1.3801 & 13.3264 & 0.4693 & 5.7342 & 1.3810\\ 
  \hline
Naive online FT& Standard &12.3065&0.4189&5.5863&1.2247&12.3169&0.4120&5.5691&1.2298\\
Online FT with meta-learning &Standard& 10.3564&0.3556&3.6403&1.1720&10.3010&0.3486&3.6287&1.1653\\
OFDA& Standard&6.2100&0.2903&2.9608&0.2899&6.1870&0.2833&2.9531&0.2745\\
OMLA& Standard&5.6230&0.2267& 2.3094&0.2680&5.4730&0.2106&2.1587&0.2541\\
  \hline
Naive online FT& Meta  &11.7650 &0.3903 &5.3825 &1.1065 &11.7302 &0.3756 &5.3530 &1.0103\\
Online FT with meta-learning & Meta&10.1024 &0.3271 &3.3195 &0.9653 &9.9366 &0.3063 &3.2987 &0.9469\\
OFDA  & Meta&5.6074 &0.2301 &2.1874 &0.2745 &5.4840 &0.2157 &2.1542 &0.9261\\ 
OMLA & Meta&\bf 5.3898 &\bf 0.2047 &\bf 2.0069 &\bf 0.2590 &\bf 5.2187 &\bf 0.1956 &\bf 1.9803 &\bf 0.2490\\ 
\bottomrule
\end{tabular}
}

\caption{Unsupervised online \textbf{monocular} depth estimation results on Eigen test scenes in the Kitti dataset. Only points with depth less than 50m are calculated.
}
\label{mono_result}
\vspace{-12pt}
\end{table*}

\section{Results in stereo matching metrics}
\label{sec:SMmetrics}
We now evaluate our approach with different network architectures according to stereo matching metrics. We use D1-all and End point Error (EPE) to compare the different approaches \cite{kitti2015}. 
Here, all the experiments are performed using the exact same protocol as in the main paper: all the online models are pretrained on Synthia dataset~\cite{synthia2016}, and the offline models are pretrained on KITTI~\cite{kitti2015} Eigen training split~\cite{eigen2015predicting}.
As shown in table ~\ref{offline_compare_stereo}, the offline methods obtain better results according to both metrics, but the naive online fine-tuning cannot bring any significant contribution to the offline method. 
Here again, the naive online learning obtains poor results in all the metrics and we observe that both \emph{OMLA}  and meta-pretraining improve significantly the performances. Even though online models report slightly lower performances than offline models, these experiments clearly illustrate the interesting potential of the online learning setting.
\begin{table}[h]

\begin{center}
\begin{tabular}{ | l  || c | c | }
\toprule
   Method  & d1-all & EPE\\
   \midrule
Godard \etal \cite{godard2017unsupervised} Offline  &  18.6883 &  2.9076  \\ 
Godard \etal \cite{godard2017unsupervised} Offline + Online & 19.3257 & 2.9803 \\
Godard \etal \cite{godard2017unsupervised} Naive & 50.2587 & 5.2140 \\
Godard \etal + \emph{OMLA}&  22.3525 &  3.5820 \\
\hline
MADNet~\cite{tonioni2019cvpr} Offline  & 17.2573 & 2.7544 \\
MADNet~\cite{tonioni2019cvpr} Offline + Online & 17.1209 & 2.7631 \\
MADNet~\cite{tonioni2019cvpr} Naive & 46.9753 & 4.9866 \\
MADNet + \emph{OMLA} &  20.2215 &  3.2014 \\
\hline
DispNet~\cite{mayer2016large} Offline  & 20.4301 & 2.9542 \\ 
DispNet~\cite{mayer2016large} Offline + Online & 20.1037 & 2.9256 \\
DispNet~\cite{mayer2016large} Naive & 51.8796 & 3.0259 \\
DispNet + \emph{OMLA} &  25.3598 &  3.3746\\

\bottomrule
\end{tabular}
\end{center}
\caption{
Comparison with different offline methods. Only points with depth less than 50m are calculated. 
}
\label{offline_compare_stereo}
\end{table}

\begin{figure*}[t]
    \centering
    \includegraphics[width=.98\textwidth]{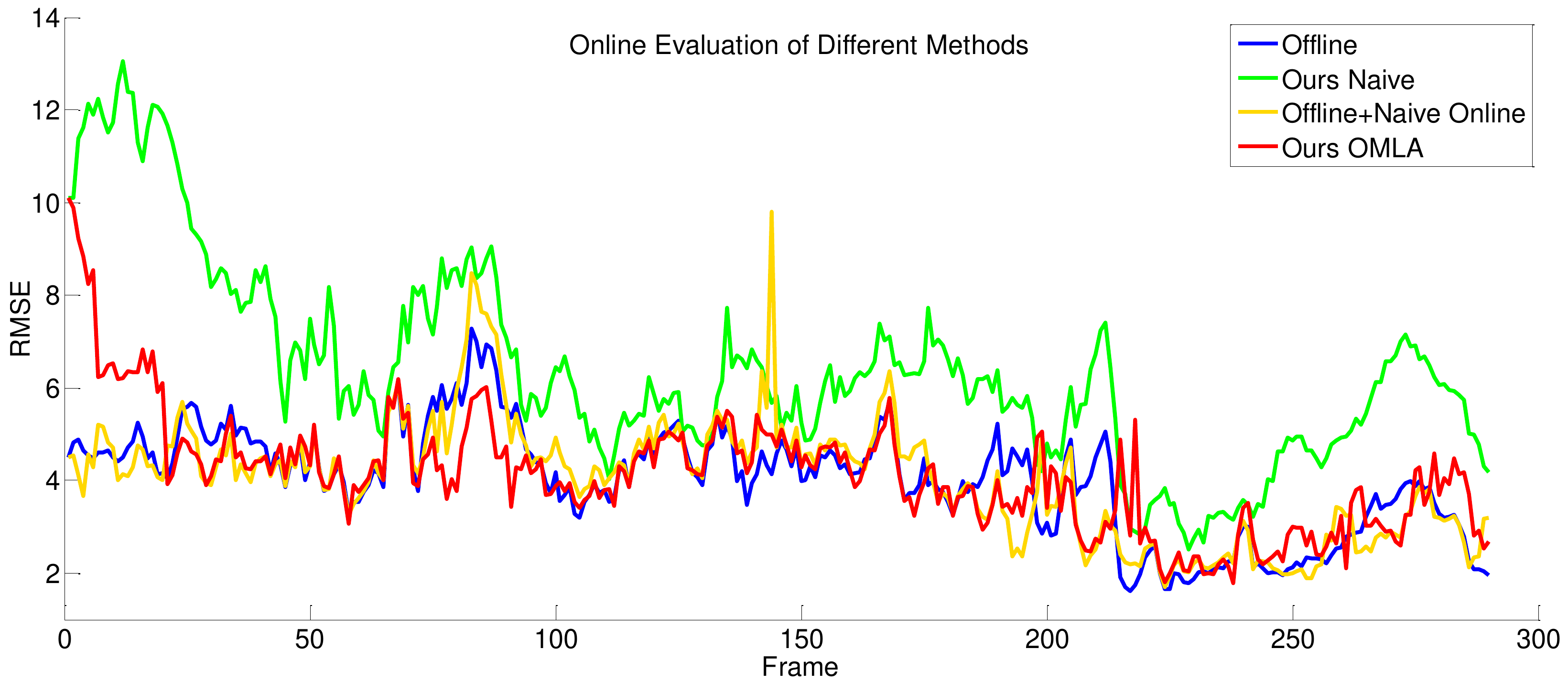}
    \caption{Online evaluation across frames of different methods on the 2011\_09\_26\_drive\_0052\_sync sequence from the KITTI Eigen test split.}
    \label{fig:supp}
\end{figure*}
\section{Illustration of Online Learning}
\label{sec:illu}
In this section we show an online evaluation over a video sequence using different methods. We select the sequence named 2011\_09\_26\_drive\_0056\_sync from KITTI Eigen testing split and perform online evaluation on it. This sequence contains only 293 frames so that online learning on such a short sequence is challenging. We illustrate the evolution of RMSE on each frame of different methods to see how each method adapt to the current environment. The offline models are trained on the KITTI Eigen training split, and online models (our \emph{OMLA} and \emph{naive}) are meta-pretrained on Synthia. All the models are based on \cite{godard2017unsupervised} and have a ResNet-50 architecture.
As shown in Fig~\ref{fig:supp}, the offline method performs well on the sequence from the beginning, since the model is trained on images visually similar to the test sequence. Applying naive online learning to the model trained offline does not improve the performance significantly, and may bring some instability to the model (e.g., at the frames around 150).
For the naive online learning model (the green line), the model performs poorly in the first frames because of the difference between the synthetic and real-world images. Then after about 20 frames, the network starts to adapt to the current environment. Note that, even if the model provides better results after 50 frames, the results are not as stable and robust as the offline methods in the following frames. The performances of our naive model are constantly worse than those of offline models.
Concerning our approach, our model with \emph{OMLA} (the red line) obtains performance competitive with the two offline models trained on the real KITTI images. Even in the first 10 frames, the model starts to adapt quickly to the new environment and shows much faster convergence than naive online learning. In the following frames, the performance of \emph{OMLA} is also more stable than with the naive approach.
Even if the sequence is rather short, our model with \emph{OMLA} can provide depth predictions with a precision similar to offline models. Such results demonstrate the effectiveness of our method.

\end{document}